\documentclass[runningheads]{llncs}

\usepackage{eccv}

\usepackage{eccvabbrv}

\usepackage{graphicx}
\usepackage{booktabs}
\usepackage{amsmath}
\usepackage{amsfonts}
\usepackage{algorithm}
\usepackage{algorithmic}
\usepackage{multirow}
\usepackage{pifont}
\usepackage[percent]{overpic}
\usepackage{xcolor}
\usepackage[absolute]{textpos}

\usepackage[accsupp]{axessibility}  % Improves PDF readability for those with disabilities.

\usepackage{hyperref}

\usepackage{orcidlink}

\begin{document}

% RPG website required watermark
\definecolor{somegray}{rgb}{0.5, 0.5, 0.5}
\newcommand{\darkgrayed}[1]{\textcolor{somegray}{#1}}
% The position is defined in absolute coords. Thus, if the page has another width,
% both numbers (4, 0.7) need to be adjusted. The argument {8} specifies the width of the textblock.
\begin{textblock}{8}(4, 0.7)
\begin{center}
\darkgrayed{This paper has been accepted for publication at the \\
European Conference on Computer Vision (ECCV), 2026}
\end{center}
\end{textblock}
% End of RPG website requried watermark

% ---------------------------------------------------------------
% TODO REVIEW: Replace with your title
\title{Low-latency Event-based Object Detection with Spatially-Sparse Linear Attention} 

% TODO REVIEW: If the paper title is too long for the running head, you can set
% an abbreviated paper title here. If not, comment out.
\titlerunning{Low-latency Event-based Object Detection with SSLA}

% TODO FINAL: Replace with your author list. 
% Include the authors' OCRID for the camera-ready version, if at all possible.
\author{
Haiqing Hao\inst{1}\orcidlink{0009-0002-0991-2009} \and
Zhipeng Sui\inst{1}\orcidlink{0000-0003-1913-1986} \and
Rong Zou\inst{2}\orcidlink{0009-0002-0434-5746} \and
Zijia Dai\inst{3}\orcidlink{0009-0007-0108-1921} \and
Nikola Zubi\'c\inst{2}\orcidlink{0000-0001-9816-2718} \and
Davide Scaramuzza\inst{2}\orcidlink{0000-0002-3831-6778} \and
Wenhui Wang\inst{1}\thanks{Corresponding author: \email{wwh@tsinghua.edu.cn}}\orcidlink{0000-0002-5884-6098} 
}

% TODO FINAL: Replace with an abbreviated list of authors.
\authorrunning{H.~Hao et al.}
% First names are abbreviated in the running head.
% If there are more than two authors, 'et al.' is used.

% TODO FINAL: Replace with your institution list.
\institute{
State Key Laboratory of Precision Measurement Technology and Instruments, Department of Precision Instrument, Tsinghua University, Beijing, China \and
Robotics and Perception Group, University of Zurich, Zurich, Switzerland \and
ShanghaiTech University, Shanghai, China
}

% \maketitle
{\let\newpage\relax\maketitle}

% fig 1

\begin{figure}[t]
     \centering
     \begin{subfigure}[b]{0.22\textwidth}
         \centering
         \includegraphics[width=\textwidth]{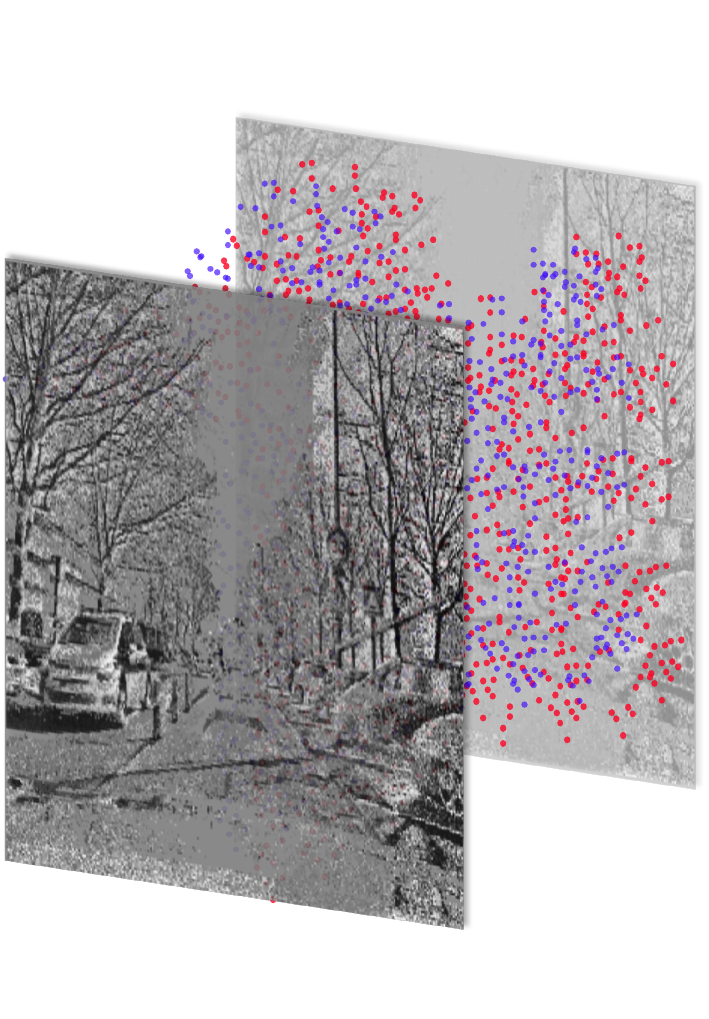}
         \caption{Events}
         \label{fig:image1}
     \end{subfigure}
     \hfill 
     \begin{subfigure}[b]{0.13\textwidth}
         \centering
         \includegraphics[width=\textwidth]{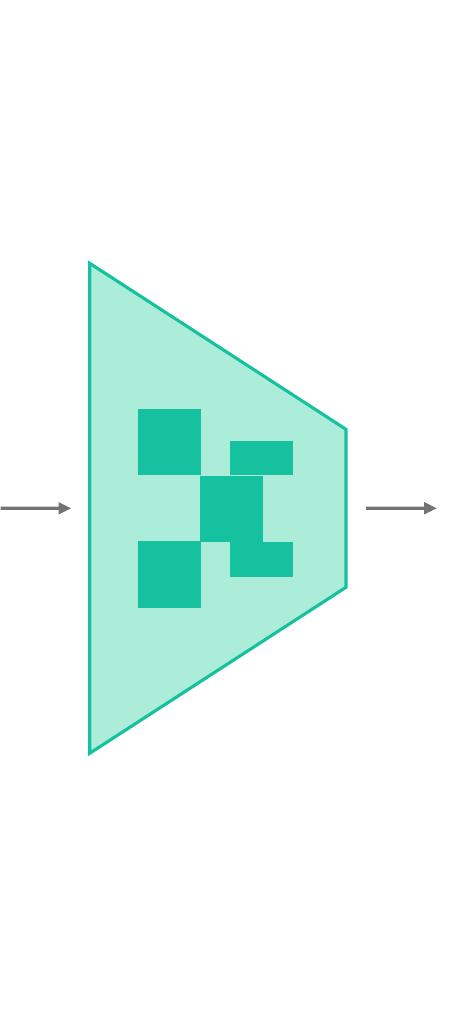}
         \caption{SSLA}
         \label{fig:image2}
     \end{subfigure}
     \hfill
     \begin{subfigure}[b]{0.18\textwidth}
         \centering
         \includegraphics[width=\textwidth]{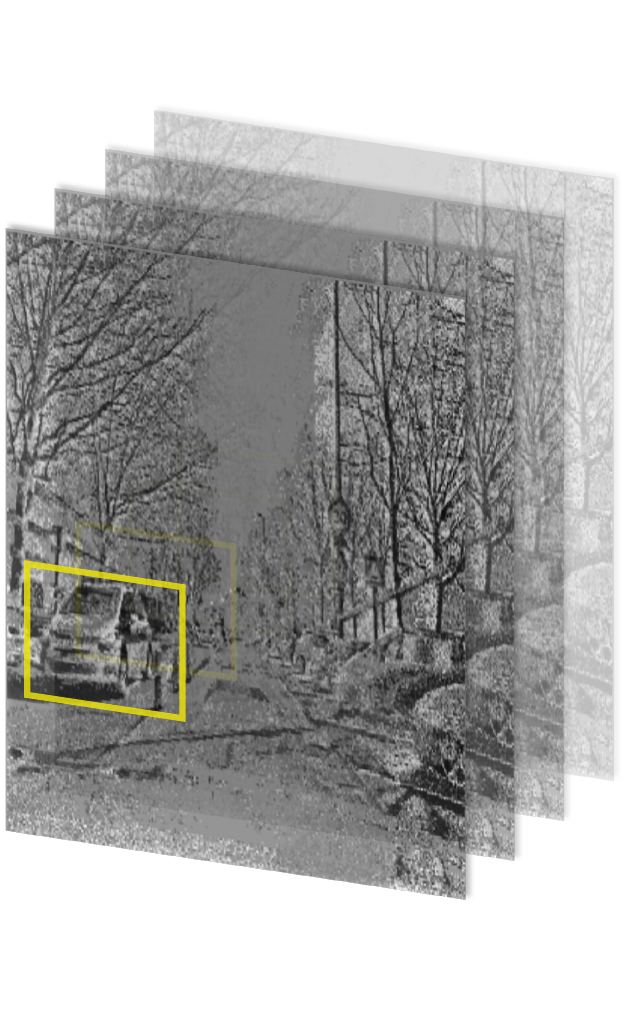}
         \caption{Detection}
         \label{fig:image3}
     \end{subfigure}
     \hfill
     \begin{subfigure}[b]{0.44\textwidth}
         \centering
         \includegraphics[width=\textwidth]{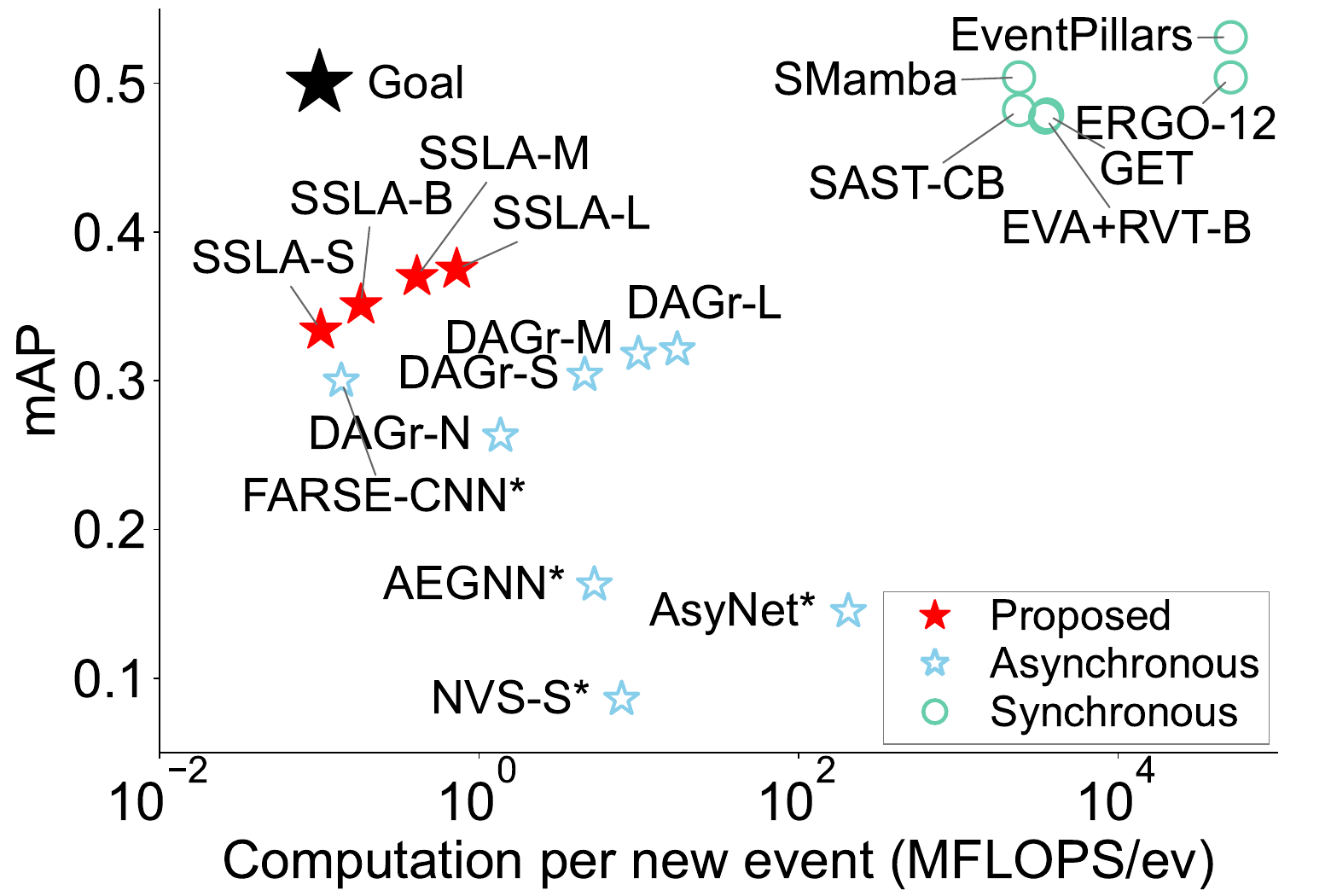}
         \caption{mAP vs. FLOPS}
         \label{fig:image4}
     \end{subfigure}
     
     \caption{Our method processes (a) asynchronous event sequence with (b) a sparsely activated linear attention neural network for (c) low-latency event-based object detection. On the Gen1 dataset, our SSLA-Det models achieve SOTA asynchronous mAP and lower FLOPS compared with previous asynchronous baselines (d). \(^{*}\) refers to AP\(_{50}\).}
     \label{fig:intro}
\end{figure}

\begin{abstract}
  Event cameras provide sequential visual data with spatial sparsity and high temporal resolution, making them attractive for low-latency object detection.
  Existing asynchronous event-based neural networks exploit this low-latency advantage by updating predictions event by event, but still suffer from two bottlenecks: recurrent architectures are difficult to train efficiently on long sequences, and improving accuracy often increases per-event computation and latency. 
  Linear attention is appealing because it enables parallel training and recurrent inference.
  However, its dense state updates make per-event computation scale with the state size, yielding a poor accuracy-efficiency trade-off for object detection, where accurate localization requires fine-grained spatial states.
  The key challenge is therefore to introduce \emph{sparse state activation} that exploits the spatial sparsity of events while preserving \emph{efficient parallel training}.
  We propose Spatially-Sparse Linear Attention (SSLA), which introduces a mixture-of-spaces state decomposition and a scatter-compute-gather training procedure, enabling state-level sparsity as well as training parallelism.
  Building on SSLA, we develop an end-to-end asynchronous linear attention model, SSLA-Det, for low-latency event-based object detection. 
  On Gen1 and N-Caltech101, SSLA-Det achieves state-of-the-art accuracy among asynchronous methods, reaching 0.375 mAP and 0.515 mAP, respectively, while reducing per-event computation by over 20\(\times\) compared with the strongest prior asynchronous baseline, demonstrating the potential of linear attention for low-latency event-based vision. Code is available at: \url{https://github.com/haohq19/ssla}.
\end{abstract}

\section{Introduction}

Event cameras provide sequential visual data with spatial sparsity and high temporal resolution, making them highly promising for low-latency perception \cite{gallego2020event, paredes2024fully, gehrig2024low}. 
Asynchronous event-based neural networks realize this potential by updating their predictions every time a new event arrives \cite{santambrogio2024farse, schaefer2022aegnn}. 
This event-driven processing paradigm is particularly appealing for object detection in latency-critical scenarios, such as autonomous driving \cite{gehrig2024low}, drone obstacle avoidance \cite{falanga2020dynamic}, and vision-based control \cite{he2024neuromorphic}.

Despite their much lower latency, existing asynchronous event-based neural networks still lag behind their synchronous counterparts in accuracy \cite{zubic2023chaos, peng2023get, zubic2024state}. 
The gap stems from two coupled architectural bottlenecks.
The first is the parallel-recurrent bottleneck: the event-by-event inference paradigm naturally relies on recurrent architectures, whereas efficient training on long event sequences requires parallelization along the sequence dimension \cite{sekikawa2019eventnet, hao2026maximizing}.
The second is the accuracy-efficiency trade-off: improving accuracy typically requires larger and deeper models, while scaling the model increases per-event computation and consequently latency.
A natural way to mitigate this trade-off is to exploit the spatial sparsity of event camera data through sparse neural network activation \cite{schaefer2022aegnn, santambrogio2024farse}. 
Nevertheless, as receptive fields expand layer by layer in deep networks, sparse inputs can still induce dense activations.
To preserve sparsity in deep layers and reduce computation, prior work has designed specialized architectures \cite{gehrig2024low, santambrogio2024farse}, but this comes with additional architectural constraints that further limit accuracy.

Linear attention\footnote{For convenience, here we use linear attention as a shorthand for parallel-trainable linear recurrent models, including state space models (SSMs) and linear recurrent neural networks (linear RNNs).} has emerged as a promising asynchronous event-based model architecture since it naturally addresses the parallel-recurrent bottleneck \cite{yang2024parallelizing, yang2024gated, peng2023rwkv, katharopoulos2020transformers, gu2024mamba}.
However, existing methods are limited to relatively simple global-level classification tasks \cite{soydan2024s7, schone2024scalable}, while more challenging local-level tasks, like object detection, remain unexplored.
The main obstacle is the poor accuracy-efficiency trade-off due to the lack of state-level sparsity, \ie, linear attention updates all elements of its state, making per-event computation scale with the state size.
This is problematic for object detection, where accurate localization requires fine-grained spatial representations and therefore a large state size \cite{hao2026maximizing}.
Although sparsifying state activation is conceptually straightforward, the key challenge is to maintain parallel training while gaining sparsity.

We address this challenge by introducing Spatially-Sparse Linear Attention (SSLA), a linear attention module with state-level sparsity while preserving its parallel training advantages.
To enable state-level sparsity, we introduce a \emph{mixture-of-spaces} (MOS) structure inspired by \cite{du2026mom} that decomposes the global state into substates with spatially overlapping receptive fields, and each event activates only a few substates based on its location.
To preserve the sparsity in deep networks, we aggregate the activations of each event from all its activated substates, preventing the expansion of the activated region.
We further propose a \emph{position-aware projection} (PAP) that projects events based on their relative positions within every activated substate, injecting state-relative spatial priors.
We derive a \emph{scatter-compute-gather} training procedure that parallelizes this sparse activation structure by sequence-level reorganization. 
Specifically, events are scattered into state-specific subsequences, computed in parallel by linear attention, and then gathered back into the original event sequence.
In this way, SSLA makes linear attention sparse in space, recurrent in time, and parallel in training.

Building on the SSLA module, we present SSLA-Det, to the best of our knowledge, the first end-to-end asynchronous linear attention model for low-latency event-based object detection (\cref{fig:intro} (a)-(c)).
Experiments on Gen1 and N-Caltech101 show that SSLA-Det achieves a substantially improved accuracy-efficiency trade-off over prior asynchronous methods (\cref{fig:intro} (d)), including state-of-the-art (SOTA) asynchronous mAP (0.375 on Gen1 and 0.515 on N-Caltech101) at much lower computational cost (over \(20\times\) reduction compared with previous SOTA \cite{gehrig2024low}).
Our contributions are as follows:
\begin{itemize}
    \item We propose an SSLA module for sequential event modeling, including a MOS structure for state-level sparsity, PAP for spatial prior encoding, and a scatter-compute-gather procedure for efficient parallel training.
    \item We present SSLA-Det, to the best of our knowledge, the first end-to-end asynchronous linear attention model for event-based object detection.
    \item SSLA-Det sets a new accuracy-efficiency frontier, reaching 0.375 mAP on Gen1 and 0.515 mAP on N-Caltech101, while reducing computation by over \(20\times\) compared with the prior asynchronous SOTA method.
\end{itemize}

\section{Related Work}

\subsection{Asynchronous Event-based Neural Networks}

Asynchronous event-based neural networks treat event camera data as different geometric structures to leverage their spatial sparsity. This strategy includes graphs \cite{li2021graph, schaefer2022aegnn, dalgaty2023hugnet, dampfhoffer2025graph, gehrig2024low}, submanifolds \cite{messikommer2020event, santambrogio2024farse}, point clouds \cite{sekikawa2019eventnet, turrero2024alerttransformer}, and sequences \cite{kamal2023associative, hao2026maximizing, schone2024scalable, soydan2024s7}.
Graph-based methods \cite{li2021graph, schaefer2022aegnn, dalgaty2023hugnet, dampfhoffer2025graph, gehrig2024low, chen2025ehgcn} transform events into sparsely connected spatial-temporal graphs, and derive local update rules on the graph for recurrent inference.
However, they have limitations in temporal accumulation \cite{dampfhoffer2025graph}, failing to handle long event sequences.
Submanifold methods \cite{messikommer2020event, santambrogio2024farse} assume that events lie on a spatial submanifold, and conduct convolution only on the submanifold to keep sparsity.
However, this submanifold assumption does not strictly hold, and thus leads to suboptimal performance.
Point cloud methods \cite{turrero2024alerttransformer, sekikawa2019eventnet} represent events as spatial-temporal 3-D point clouds, and process with PointNet \cite{qi2017pointnet}.
This approach is limited to shallow neural networks, and thus has difficulty in challenging tasks.
Sequence-based methods use causal sequence-to-sequence models for event processing including softmax attention or linear recurrent models.
For example, \cite{kamal2023associative} uses attention to process batched events at each timestep.
EventSSM \cite{schone2024scalable} and S7 \cite{soydan2024s7} use SSMs, which enable parallel training and recurrent inference, but are limited to global-level classification tasks.
EVA \cite{hao2026maximizing} explores linear attention for event-based object detection but still demands a dense backbone, while our method is fully end-to-end asynchronous.

\subsection{State-level Sparsity in Linear Attention}

A recent direction to improve linear attention is to introduce state-level sparsity.
Mixture-of-Memories \cite{du2026mom} and Sparse State Expansion \cite{pan2026scaling} maintain multiple independent states and use a learned router to send each token to only a few memories.
Our SSLA follows the same sparse state activation idea but is fundamentally different in that the construction of substates is spatially structured, which enables geometric routing of event embeddings and admits the position-aware projection that encodes spatial inductive bias.
Concurrent work \cite{sekikawa2026col2a} also introduces local states in linear attention for local-level event-based vision, but requires spatial contraction at discrete timestamps, which is not asynchronously, event-by-event trainable. 
Our work, instead, uses a scatter-compute-gather algorithm to reorganize events into subsequences and does not rely on spatial contraction, which is fully event-by-event asynchronous.

\section{Method}

\subsection{Problem Formulation: Asynchronous Event Processing}

Event camera data are represented as a sequence of events \(\mathcal{E}=\{e_i\}_{i=1}^L\), where each event \(e_i = (\mathbf{x}_i, t_i, p_i)\) carries its spatial coordinates \(\mathbf{x}_i \in \mathbb{R}^2\), a timestamp \(t_i\in\mathbb{R}\) and a polarity \(p_i \in \{+1, -1\}\). 
The sequence is temporally ordered so that \(t_i \leq t_{i+1}\).
Asynchronous event processing learns a causal stateful neural network \(\mathcal{M}\) that takes \(\mathcal{E}\) as input and makes predictions incrementally, which can be formulated as a causal sequence-to-sequence problem from \(\{e_i\}_{i=1}^L\) to \(\{\hat{\mathbf{y}}_i\}_{i=1}^L\).
Specifically, for each new incoming event \(e_i\), the model updates its state \(\mathbf{S}_i\) and produces a new prediction \(\hat{\mathbf{y}}_i\), as
\((\hat{\mathbf{y}}_i, \mathbf{S}_i) = \mathcal{M}(e_i, \mathbf{S}_{i-1})\).

\subsection{Preliminaries: Linear Attention}
Linear attention models (linear RNNs, SSMs) are linear-time alternatives to softmax attention \cite{vaswani2017attention} for sequence-to-sequence modeling.
Causal linear attention has an equivalent parallel and recurrent form, which enables both parallel training and recurrent inference \cite{katharopoulos2020transformers}.
Given an input sequence of embeddings \(\{\mathbf{z}_i\}_{i=1}^L\), a causal linear attention module updates its hidden state \(\mathbf{S}_i\) and computes outputs \(\{\mathbf{o}_i\}_{i=1}^L\) as
\begin{equation}
\label{eq:recurrent-linear-attention}
\begin{aligned}
    \mathbf{S}_i &= g(\mathbf{z}_i) \odot \mathbf{S}_{i-1} + \phi(\mathbf{z}_i), \\
    \mathbf{o}_i &= \rho(\mathbf{z}_i, \mathbf{S}_i),
\end{aligned}
\end{equation}
where \(g, \phi, \rho\) are learnable projections (gating, updating, and output) and \(\odot\) is the Hadamard product.
This recurrent linear attention is parallel trainable with parallel scan \cite{gu2024mamba} or chunk-wise algorithms \cite{yang2024gated}, which is training-efficient on long event sequences.
For convenience, we denote the map of linear attention as 
\(\textbf{LinearAttention}: \{\mathbf{z}_i\}_{i=1}^L\mapsto {\{\mathbf{o}_i\}_{i=1}^L}\) in this paper.

\subsection{Spatially-Sparse Linear Attention for Event Sequence Modeling}

\begin{figure}[t]
    \centering
    \includegraphics[width=1.0\linewidth]{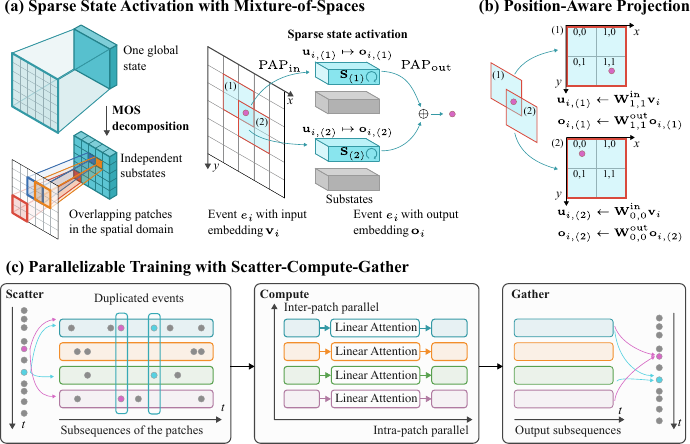}
    \caption{\textbf{Overview of the Spatially-Sparse Linear Attention (SSLA) module}. 
    (a) The global state is decomposed into independent substates, one per overlapping patch in the spatial domain.
    (1) and (2) are \(2\times 2\) patch examples that cover the event \(e_i\) (for brevity, we omit the other patches covering \(e_i\)). Their states are updated with the embedding of \(e_i\), and the output embedding of \(e_i\) is the summation of the interim outputs from all the patches covering \(e_i\).
    (b) In each patch, the embeddings are projected based on their relative position within the patch.
    (c) During training, with an event sequence as input, the events are scattered into patch-wise subsequences (duplicating each event across all the patches covering it), computed in parallel by a weight-shared linear attention, then gathered back to the original event order.
    } 
    \label{fig:ssla-overview}
\end{figure}

We introduce the Spatially-Sparse Linear Attention (SSLA) module for asynchronous event processing, which is illustrated in \cref{fig:ssla-overview}.
The SSLA module takes an event sequence \(\mathcal{E}\) with embeddings \(\mathbf{v}_i \in\mathbb{R}^{D_{in}}\) as input, and outputs \(\mathcal{E}\) with updated embeddings \(\mathbf{o}_i\in\mathbb{R}^{D_{out}}\).
To exploit the spatial sparsity of events, we first introduce a \emph{mixture-of-spaces} (MOS) decomposition of the hidden state, which enables state-level spatial sparsity.
We incorporate a \emph{position-aware projection} in the SSLA module to encode spatial priors into event embeddings.
To preserve the training efficiency of linear attention, we derive a \emph{scatter-compute-gather} procedure for parallel training by reorganizing events to independent subsequences.

% Figure 2

\subsubsection{Sparse State Activation with Mixture-of-Spaces.}
We define \(\Omega\) as the spatial domain of the event camera data, and decompose it into spatially local and overlapping patches, with each patch maintaining its own state independently.
We construct these patches by applying a sliding window of size \(P \times P\) with a stride of 1 over \(\Omega\).
Let \(\mathcal{P} = \{1, \dots, K\}\) denote the indices of these patches, where each patch \(k \in \mathcal{P}\) covering a spatial region \(\mathcal{R}_k \subset \Omega\) maintains state \(\mathbf{S}_k\).

For \(e_i\) at \(\mathbf{x}_i\), we activate only specific patches that contain \(\mathbf{x}_i\) for sparse state activation.
We define the set of active patches of \(e_i\) as
\begin{equation}
    \mathcal{K}_i = \{k \in \mathcal{P} \mid \mathbf{x}_i \in \mathcal{R}_k\}.
\end{equation}
We pad the image domain to ensure that each event activates a constant number of states, \ie, \(|\mathcal{K}_i| = P^2 \triangleq A \).
These \(A\) states are updated by \cref{eq:recurrent-linear-attention} with event embeddings, and generate interim outputs \(\{\mathbf{o}_{i,k} \mid k\in \mathcal{K}_i\}\).
All active patches share the same linear attention parameters but maintain independent states.
We aggregate the interim outputs as the updated embedding from all activated patches:
\begin{equation}
    \label{eq:aggregate-no-pap}
    \mathbf{o}_i = \sum_{k \in \mathcal{K}_i} \mathbf{o}_{i,k}.
\end{equation}
After aggregation, the embedding sequence length is unchanged, which preserves sparsity in deep layers.

\subsubsection{Position-Aware Projection.}

Sharing identical embeddings \(\mathbf{v}_i\) for one event in all its activated patches ignores the spatial prior of the event, since the event is located at different relative positions in different patches.
We therefore introduce a position-aware projection (PAP) of the embedding, which projects the embedding based on the event's relative positions inside the patches.

For an event at \(\mathbf{x}_i\) in patch \(k\) with top-left global coordinates \(\mathbf{c}_k \in \Omega\), we compute its relative position in the patch as
\begin{equation}
    \boldsymbol{\delta}_{i,k} = \mathbf{x}_i - \mathbf{c}_k, \quad \text{where } \boldsymbol{\delta}_{i,k} \in \{0,..., P-1\}^2.
\end{equation}
We define the PAP as a linear transform by \(\mathbf{W}^{\text{in}}\in \mathbb{R}^{P\times P\times D_{out}\times D_{in}}\) and \(\mathbf{W}^{\text{out}}\in \mathbb{R}^{P\times P \times D_{out}\times D_{out}}\). 
The input embedding \(\mathbf{v}_i\) is projected by
\begin{equation}
    \mathbf{u}_{i,k} \in \mathbb{R}^{D_{out}}\leftarrow  \mathbf{W}^{\text{in}}[\boldsymbol{\delta}_{i,k}]\mathbf{v}_i.
\end{equation}
Then we use \(\mathbf{u}_{i,k}\) in patch \(k\) as input to linear attention. 
Similarly, before aggregating the interim outputs, we also conduct a PAP, and \cref{eq:aggregate-no-pap} becomes
\begin{equation}
    \label{eq:aggregate}
    \mathbf{o}_i = \sum_{k \in \mathcal{K}_i}  \mathbf{W}^{\text{out}}[\boldsymbol{\delta}_{i,k}] \mathbf{o}_{i,k}.
\end{equation}
This projection encodes spatial priors with learnable parameters, expanding the capability of the model to capture spatial patterns and information.
\begin{algorithm}[t]
\caption{Spatially-Sparse Linear Attention (SSLA) Module  Training}
\label{alg:ssla}
\begin{algorithmic}[1]
\REQUIRE Events \(\mathcal{E}\), embeddings \(\{\mathbf{v}_i\}_{i=1}^{L}\), coordinates \(\{\mathbf{x}_i\}_{i=1}^L\), Patches \(\mathcal{P}\), lookup table \(T:\mathbf{x}\in\Omega \mapsto \{\left(k, \delta\right)\}\).
\ENSURE Updated embeddings \(\{\mathbf{o}_i\}_{i=1}^L\) in the same order as the input.
\STATE  Initialize an empty embedding sequence \(\mathcal{U}\) of length \(AL\).

\FOR{each event \(e_i\) \textbf{in parallel}}
    \STATE Lookup \(\mathbf{x}_i\): active patch indices \(\{k\mid k\in \mathcal{K}_i\}\), relative positions \(\{\boldsymbol{\delta}_{i,k}\mid k\in \mathcal{K}_i\}\);
    \STATE Position-aware projection: \(\mathbf{u}_{i,k} \leftarrow  \mathbf{W}^{\text{in}}[\boldsymbol{\delta}_{i,k}]\mathbf{v}_i\);
    \STATE Assign \(\mathcal{U}[A(i-1):Ai]\leftarrow \{\mathbf{u}_{i,k}\mid k \in \mathcal{K}_i\}\) 
\ENDFOR

\STATE Scatter: stable-sort \(\mathcal{U}\) based on \(k\), and cache the permutation \(\pi\);
\STATE Split \(\mathcal{U}\) to patch-specific subsequences \(\{\mathcal{U}_k\mid k \in \mathcal{P}\}\); 

\FOR{each patch \(k \in \mathcal{P}\) \textbf{in parallel}}
    \STATE \(\mathcal{O}_k \leftarrow \textbf{LinearAttention}(\mathcal{U}_k)\) using \cref{eq:recurrent-linear-attention};
\ENDFOR

\STATE Concat  \(\mathcal{O}_k\) into interim output sequence \(\mathcal{O}\);
\STATE Gather: \(\mathcal{O}\leftarrow \mathcal{O}[\pi^{-1}]\). We get \(\mathcal{O}[A(i-1):Ai] = \{\mathbf{o}_{i,k}\mid k \in \mathcal{K}_i\}\);
\FOR{each event \(e_i\) \textbf{in parallel}}
    \STATE Position-aware projection: \(\mathbf{o}_{i,k} \leftarrow \mathbf{W}^{\text{out}}[\boldsymbol{\delta}_{i,k}]\mathbf{o}_{i,k}\);
\ENDFOR

\STATE Aggregate: \(\mathbf{o}_i = \sum\mathcal{O}[A(i-1):Ai]\);

\RETURN $\{\mathbf{o}_i\}_{i=1}^L$
\end{algorithmic}
\end{algorithm}

\subsubsection{Parallelizable Training with Scatter-Compute-Gather.}
While the MOS structure enables state-level sparsity, it breaks the single state form of linear attention, and thus poses challenges in efficient parallel training on GPUs.
To address this, we derive a \emph{scatter-compute-gather} algorithm, which allows both intra-patch (temporal) and inter-patch (spatial) parallelism to speed up training.

\paragraph{Scatter.}
Let \(\mathbf{u}_{i,k}\) denote the projected embeddings of event \(e_i\) for the active patch \(k \in \mathcal{K}_i\).
We first reorganize the input event sequence \(\mathcal{E}\) to \(K\) patch-specific subsequences, by constructing a subsequence for each patch \(k\) as
\begin{equation}
    \mathcal{U}_k = \{ \mathbf{u}_{i,k} \mid i \text{ such that } k \in \mathcal{K}_i \}, \quad \forall k \in \mathcal{P},
\end{equation}
where \(\mathcal{U}_k\) is ordered by \(t_i\).
We implement this with a precomputed lookup table that maps each coordinate in \(\Omega\) to the indices of the patches covering it, together with its relative position inside each patch.
Using the table, we expand \(\mathcal{E}\) to a projected sequence \(\mathcal{U}\) of length \(AL\), where each event contributes \(A\) consecutive projected embeddings, one for each of its activated patches.
We apply stable sorting on  \(\mathcal{U}\) based on the patch indices of each element, forming a reorganized sequence with embeddings in the same patch grouped together as the subsequence, while preserving the temporal order within each \(\mathcal{U}_k\).
The resulting permutation is cached and later reused in the gather step.

\paragraph{Compute.}
All patches share the same parameters but maintain independent states, so that the \(K\) subsequences can be computed in parallel, achieving inter-patch parallelism.
We apply linear attention (\cref{eq:recurrent-linear-attention}) to each subsequence \(\mathcal{U}_k\)
\begin{equation}
    \mathcal{O}_k = \textbf{LinearAttention}(\mathcal{U}_k),
\end{equation}
where \(\mathcal{O}_k\) contains the interim outputs for events contained in patch \(k\).
In each patch, the linear attention is standard, which achieves intra-patch parallelism.

\paragraph{Gather.}
Finally, we restore the outputs to the original expanded event sequence order using the cached permutation and aggregate the interim outputs from all active patches of each event.
Specifically, we first apply the PAP to each interim output \(\mathbf{o}_{i,k}\), and then sum them over \(k \in \mathcal{K}_i\) according to \cref{eq:aggregate}.
This gather step is efficient because it only involves indexed reordering and reduction. 
The training procedure of the SSLA module is summarized in \cref{alg:ssla}.

\subsection{SSLA-Det Architecture}

\begin{figure}[t]
    \centering
    \includegraphics[width=1.0\linewidth]{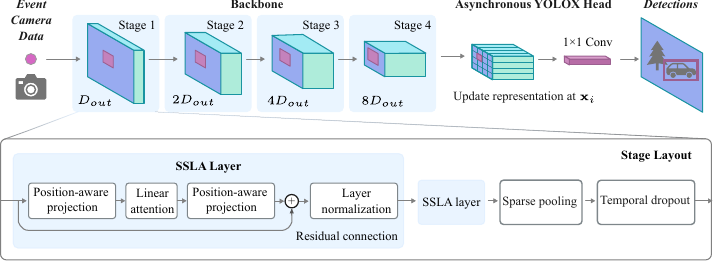}
    \caption{\textbf{Overview of the SSLA-Det model.} 
    \textbf{Top:} The \emph{fully asynchronous} event-based object detector. 
    Events are processed by a 4-stage asynchronous backbone, and each stage doubles the output embedding dimension. 
    Each output embedding updates the representation of its position, and an asynchronous YOLOX head gives the detections. 
    The red region marks the area sparsely activated by an event.
    \textbf{Bottom:} Stage layout.
    Each stage has 2 SSLA layers followed by sparse pooling and temporal dropout (only SSLA layers at stage 4).
    In one SSLA layer, event embeddings are processed by the SSLA module, including two position-aware projections and a patch-wise linear attention.
    A residual connection and a layer normalization are used for training stability.
    }
    \label{fig:ssla-detail}
\end{figure}

We propose the SSLA-Det model for asynchronous event-based object detection.
An overview of our neural network is shown in \cref{fig:ssla-detail}, which consists of an asynchronous backbone and a YOLOX detection head \cite{ge2021yolox}.
The backbone has 4 stages, each containing 2 SSLA module layers.
In the SSLA layers, we use residual connections \cite{he2016deep} and layer normalization \cite{ba2016layer} to stabilize training.
In the first 3 stages, we use one sparse pooling and one temporal dropout layer introduced in \cite{santambrogio2024farse}, which compresses event sequence to reduce computation while preserving the high temporal resolution of events.
All of the above layers are asynchronous, which makes the backbone fully asynchronous.

SSLA module is agnostic to the specific design of linear attention mechanism, allowing for the integration of any variant, including linear RNNs and SSMs.
We use a real-valued Linear Recurrent Unit \cite{orvieto2023resurrecting} implemented by Triton \cite{tillet2019triton} in our model for hardware efficiency.
The embedding dimension \(D_{out}\) has an expansion of 2\(\times\) in each stage.

For each step, the input to the model is a raw event, with polarity and time difference \(\mathbf{v}_i = \left[p_i, \mathrm{\Delta} t_i\right] \in \mathbb{R}^2\) as the embedding, and the backbone generates \(\mathbf{o}_i\).
We form a spatially fine-grained representation \(\mathbf{R}\in\mathbb{R}^{H_{out}\times W_{out}\times D_{out}}\) from the asynchronous output, by updating \(\mathbf{o}_i\) to \(\mathbf{R}[\mathbf{x}_i]\).
To achieve an end-to-end asynchronicity, we modify the YOLOX head by changing all the convolutions to \(1\times 1\).
Each backbone output only updates the head predictions at position \(\mathbf{x}_i\), making the YOLOX head also asynchronous.
The whole SSLA-Det model is therefore end-to-end fully asynchronous, leading to minimal latency.

% Figure 3

\section{Experiments}
\label{sec:experiments}

\subsection{Experimental Setup}

\subsubsection{Datasets.} 
Following previous work \cite{gehrig2024low}, we evaluate on the N-Caltech101 Detection \cite{orchard2015converting} and the Gen1 Detection \cite{de2020large} datasets.
N-Caltech101 consists of recordings captured by a DAVIS240 event camera with a resolution of \(240 \times 180\) pixels, undergoing saccadic motion in front of a projector displaying Caltech101 images.
Bounding box annotations were manually added in post-processing, with 101 classes.
Gen1 is a more challenging, large-scale benchmark for automotive scenarios, recorded by an ATIS event camera with a resolution of \(304 \times 240\) pixels.
The dataset has two categories of 228,123 cars and 27,658 pedestrians.
Following previous works \cite{gehrig2024low, perot2020learning}, we filter out bounding boxes with a diagonal below 30 pixels or width below 20 pixels in Gen1.

\subsubsection{Training Details.}
All experiments were conducted with PyTorch 2.6.0 \cite{paszke2019pytorch} on NVIDIA Ampere GPUs (A800/RTX 3090).
We design four variants of our SSLA-Det model: small (SSLA-S), base (SSLA-B), medium (SSLA-M), and large (SSLA-L), by scaling the embedding dimension \(D_{out}\) of the first stage to 12, 16, 24 and 32.
We use AdamW \cite{loshchilov2018decoupled} optimizer.
For Gen1, we train 40 epochs using a batch size of 32 and a base learning rate of \(1 \times 10^{-3}\) with a cosine decay.
We use random flipping with probability 0.5 and random dropout of input events with the keep ratio sampled from \(\mathcal{U}(0.8, 1.0) \).
For N-Caltech101, we train 200 epochs with a batch size of 64. 
In addition to random dropout, we apply random cropping to 75\(\%\) of the full resolution with probability 0.2 and random translation by up to 10\(\%\) of the full resolution following the implementations of \cite{gehrig2024low}. 
We also use exponential model averaging \cite{izmailov2018averaging}.

\subsection{Results}

\subsubsection{Gen1 Automotive.}

We compare our SSLA-Det models with asynchronous event-based detection baselines \cite{li2021graph, messikommer2020event, schaefer2022aegnn, santambrogio2024farse, gehrig2024low}, and use synchronous methods as reference \cite{hao2026maximizing, zubic2023chaos, fan2025eventpillars, peng2023get, yang2025smamba, peng2024scene}.
The performance is evaluated in accuracy with mean average precision (mAP) \cite{lin2014microsoft} and efficiency with average floating point operations (FLOPS) for every new event.
We observed that some prior works report AP\(_{50}\) whereas others report mAP, and these values are sometimes presented together in the literature. 
To avoid potentially misleading comparisons, we separate them in \cref{tab:gen1-results}.

% Table 1
\begin{table}[t]
\caption{\textbf{Object detection results on the Gen1 Detection dataset.} Async. refers to asynchronous methods.}
\label{tab:gen1-results}
\centering
\begin{tabular}{lcccc}
\toprule
Method & Async. & mAP\((\uparrow)\) & AP\(_{50}\)\((\uparrow)\) & MFLOPS/ev\((\downarrow)\) \\
\midrule
EVA+RVT-B \cite{hao2026maximizing} & \textcolor{red}{\ding{55}} & 0.477 & -     & 3.5 \(\times10^3\)  \\
GET \cite{peng2023get} & \textcolor{red}{\ding{55}} & 0.479 & -     & 3.6 \(\times10^3\) \\
SAST-CB \cite{peng2024scene} & \textcolor{red}{\ding{55}} & 0.482 & -     & 2.4 \(\times10^3\) \\
SMamba \cite{yang2025smamba} & \textcolor{red}{\ding{55}} & 0.504 & -     & 2.4 \(\times10^3\) \\
ERGO-12 \cite{zubic2023chaos} & \textcolor{red}{\ding{55}} & 0.504 & -     & 50.8 \(\times10^3\) \\
EventPillars \cite{fan2025eventpillars} & \textcolor{red}{\ding{55}} & 0.531 & -     & 50.8 \(\times10^3\)  \\
NVS-S \cite{li2021graph} & \textcolor{green}{\checkmark} & -     & 0.086 & 7.80  \\
AsyNet \cite{messikommer2020event} & \textcolor{green}{\checkmark} & -     & 0.145 & 205 \\
AEGNN \cite{schaefer2022aegnn} & \textcolor{green}{\checkmark} & -     & 0.163 & 5.26  \\
FARSE-CNN \cite{santambrogio2024farse} & \textcolor{green}{\checkmark} & -     & 0.300 & 0.137 \\
DAGr-N \cite{gehrig2024low} & \textcolor{green}{\checkmark} & 0.263 & -     & 1.36  \\
DAGr-S \cite{gehrig2024low} & \textcolor{green}{\checkmark} & 0.304 & -     & 4.58  \\
DAGr-M \cite{gehrig2024low} & \textcolor{green}{\checkmark} & 0.318 & -     & 9.94  \\
DAGr-L \cite{gehrig2024low} & \textcolor{green}{\checkmark} & 0.321 & -     & 17.4  \\
\midrule
SSLA-S (Ours) & \textcolor{green}{\checkmark} & 0.334          & 0.629          & \textbf{0.102} \\
SSLA-B (Ours) & \textcolor{green}{\checkmark} & 0.351          & 0.655          & 0.182          \\
SSLA-M (Ours) & \textcolor{green}{\checkmark} & 0.370          & 0.670          & 0.408          \\
SSLA-L (Ours) & \textcolor{green}{\checkmark} & \textbf{0.375} & \textbf{0.675} & 0.724          \\
\bottomrule
\end{tabular}
\end{table}

Our SSLA-Det models consistently improve the accuracy-efficiency trade-off over existing asynchronous baselines.
Notably, compared to the strongest prior asynchronous baseline DAGr-L \cite{gehrig2024low}, our smallest model SSLA-S achieves higher mAP (0.334 vs. 0.321) while reducing the computational cost by about 171\(\times\) (0.102 vs. 17.4 MFLOPS/ev).
Our largest model, SSLA-L, achieves an mAP of 0.375, setting a new SOTA for asynchronous detection on Gen1, with \(>20\times\) reduction of FLOPS to previous best model DAGr-L (0.724 M/ev vs. 17.4 M/ev).

\cref{fig:visualization-gen1} visualizes the detection results on Gen1. 
\cref{fig:visualization-gen1} (a)-(d) and \cref{fig:visualization-gen1} (e)-(h) show the detected cars and pedestrians, respectively.
\cref{fig:visualization-gen1} (i)-(l) provide qualitative understandings of the typical failure cases.
Specifically, \cref{fig:visualization-gen1} (i) and (j) show the false negatives, mainly caused by the lack of relative motion between the event camera and the target, which leads to missing event data.
\cref{fig:visualization-gen1} (k) and (l) show the false positives caused by missing annotations. 

% fig 4
\newcommand{\imgwithlabel}[3]{%
  \begin{overpic}[width=#1]{#2}
    \put(2,68){\color{white}\bfseries #3} % 左上角，坐标可调
  \end{overpic}%
}

\begin{figure*}[t]
  \centering
  \setlength{\tabcolsep}{1.5pt}
  \renewcommand{\arraystretch}{0.9}
  \begin{tabular}{cccc}
    \begin{tabular}{@{}c@{}}
      \imgwithlabel{0.24\textwidth}{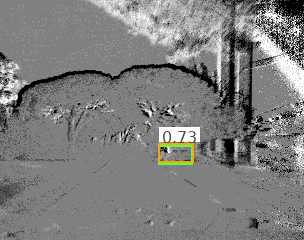}{(a)} \\
    \end{tabular} &
    \begin{tabular}{@{}c@{}}
      \imgwithlabel{0.24\textwidth}{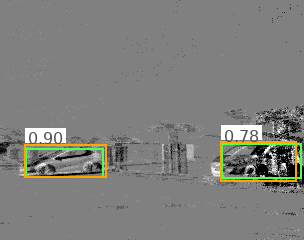}{(b)} \\
    \end{tabular} &
    \begin{tabular}{@{}c@{}}
      \imgwithlabel{0.24\textwidth}{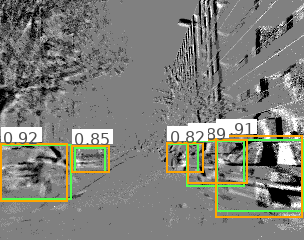}{(c)} \\
    \end{tabular} &
    \begin{tabular}{@{}c@{}}
      \imgwithlabel{0.24\textwidth}{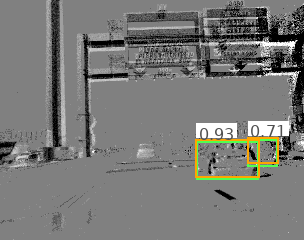}{(d)} \\
    \end{tabular} \\

    \begin{tabular}{@{}c@{}}
      \imgwithlabel{0.24\textwidth}{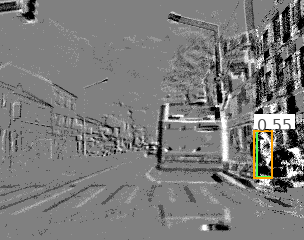}{(e)} \\
    \end{tabular} &
    \begin{tabular}{@{}c@{}}
      \imgwithlabel{0.24\textwidth}{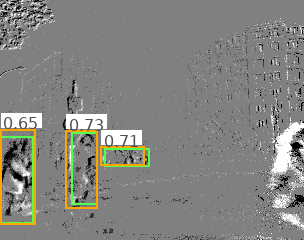}{(f)} \\
    \end{tabular} &
    \begin{tabular}{@{}c@{}}
      \imgwithlabel{0.24\textwidth}{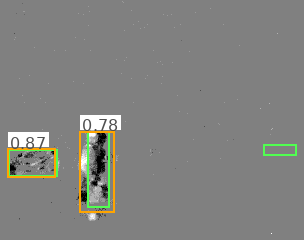}{(g)} \\
    \end{tabular} &
    \begin{tabular}{@{}c@{}}
      \imgwithlabel{0.24\textwidth}{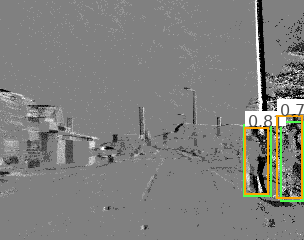}{(h)} \\
    \end{tabular} \\

    \begin{tabular}{@{}c@{}}
      \imgwithlabel{0.24\textwidth}{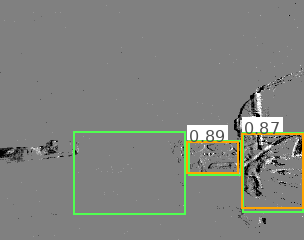}{(i)} \\
    \end{tabular} &
    \begin{tabular}{@{}c@{}}
      \imgwithlabel{0.24\textwidth}{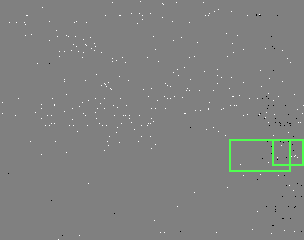}{(j)} \\
    \end{tabular} &
    \begin{tabular}{@{}c@{}}
      \imgwithlabel{0.24\textwidth}{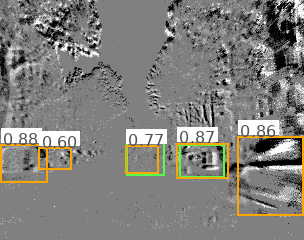}{(k)} \\
    \end{tabular} &
    \begin{tabular}{@{}c@{}}
      \imgwithlabel{0.24\textwidth}{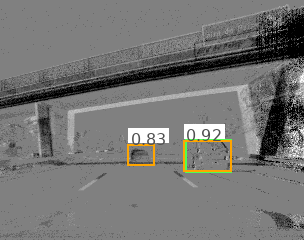}{(l)} \\
    \end{tabular}
  \end{tabular}
  \caption{\textbf{Visualization of the detection results on the Gen1 dataset.} Green boxes denote ground truth and orange boxes denote predictions with confidence scores. Predicted boxes with confidence scores below 0.5 are removed. (a)-(d): Cars. (e)-(h): Pedestrians. (i)-(l): Failure cases. }
  \label{fig:visualization-gen1}
\end{figure*}

% Table 2
\begin{table}[htbp]
\centering
\caption{\textbf{Object detection results on the N-Caltech101 Detection dataset.} Async. refers to asynchronous methods.} 
\label{tab:ncaltech-results}
\begin{tabular}{lcccc}
\toprule
Method & Async. & mAP\((\uparrow)\) & AP\(_{50}(\uparrow)\) & MFLOPS/ev\((\downarrow)\)\\
\midrule
NVS-S \cite{li2021graph} & \textcolor{green}{\checkmark} & - & 0.346 & 7.80  \\
AsyNet \cite{messikommer2020event} & \textcolor{green}{\checkmark} & - & 0.643 & 200 \\
AEGNN \cite{schaefer2022aegnn} & \textcolor{green}{\checkmark} & - & 0.595 & 7.41  \\
EHGCN \cite{chen2025ehgcn} & \textcolor{green}{\checkmark} & - & 0.694 & 1.06  \\
DAGr-N \cite{gehrig2024low} & \textcolor{green}{\checkmark} & - & 0.629 & 2.28  \\
DAGr-S \cite{gehrig2024low} & \textcolor{green}{\checkmark} & - & 0.702 & 6.85  \\
DAGr-M \cite{gehrig2024low} & \textcolor{green}{\checkmark} & - & 0.727 & 12.2  \\
DAGr-L \cite{gehrig2024low} & \textcolor{green}{\checkmark} & - & 0.732 & 18.9  \\
\midrule
SSLA-S (Ours) & \textcolor{green}{\checkmark} & 0.444          & 0.681          & \textbf{0.131} \\
SSLA-B (Ours) & \textcolor{green}{\checkmark} & 0.483          & 0.720          & 0.233          \\
SSLA-M (Ours) & \textcolor{green}{\checkmark} & 0.495          & 0.724          & 0.522          \\
SSLA-L (Ours) & \textcolor{green}{\checkmark} & \textbf{0.515} & \textbf{0.743} & 0.926          \\
\bottomrule
\end{tabular}
\end{table}

\subsubsection{N-Caltech101.}
\cref{tab:ncaltech-results} presents the performance of our models on the N-Caltech101 dataset.
Consistent with the Gen1 results, our models achieve a superior accuracy-efficiency trade-off among asynchronous methods.
In particular, SSLA-L reaches an AP\(_{50}\) of 0.743 with a computational cost of only 0.926 MFLOPS/ev.
Compared with the previous best asynchronous baseline DAGr-L, SSLA-L improves AP\(_{50}\) by 1.1 points (0.743 vs.\ 0.732) while using \(20\times\) fewer MFLOPS per event (0.926 vs.\ 18.9).

\subsection{Timing Experiments}

\subsubsection{Training Efficiency.}

We show the training efficiency benefit of linear attention with sequential parallelism.
We replace linear attention with a long short-term memory (LSTM) baseline \cite{hochreiter1997long} with the same hidden dimension as SSLA-S.
\cref{tab:timing-train} compares SSLA with an LSTM baseline from the official PyTorch implementation under the same training setup on Gen1.
We compare the training time per epoch, which is measured on 4 NVIDIA A800 GPUs.
SSLA-S reduces the epoch time from 1.05 to 0.25 hours (4.2\(\times\)), but at the cost of a drop in mAP (from 0.353 to 0.334), mainly caused by a lower FLOPS.
At a similar FLOPS level, SSLA-B achieves a comparable mAP to LSTM (0.351 vs.\ 0.353) and higher AP\(_{50}\) (0.655 vs.\ 0.631), while reducing train time from 1.05 to 0.28 hours (3.8\(\times\)).

%Table 3
\begin{table}[htbp]
\centering
\caption{\textbf{Comparison of training efficiency between SSLA and an LSTM baseline on Gen1.} Training time per epoch is measured on 4 NVIDIA A800 GPUs.}
\label{tab:timing-train}
\begin{tabular}{lccccc}
\toprule
Method & Time/epoch (h)\((\downarrow)\) & mAP\((\uparrow)\) & AP\(_{50}\)\((\uparrow)\)  & Params (M) & MFLOPS/ev\((\downarrow)\) \\
\midrule
SSLA-S & \textbf{0.25} & 0.334 & 0.629 & 0.508 & \textbf{0.102} \\
SSLA-B & 0.28 & \textbf{0.351} & \textbf{0.655} & 0.902 & 0.182 \\
LSTM   & 1.05 & \textbf{0.353} & 0.631 & 0.606 & 0.176 \\
\bottomrule
\end{tabular}
\end{table}

\subsubsection{Inference Latency.}

We measure the latency of SSLA-Det as the time required to process one newly arrived event in a recurrent, event-by-event setting.
We implement a recurrent C++ version of SSLA-Det and benchmark the latency on a single core of an AMD Ryzen 9 9950X3D CPU.
As shown in \cref{tab:timing-inference}, our models achieve a low latency of less than \(10\,\mu s\), which is lower than the sensor transmission latency of approximately \(200\,\mu s\) \cite{gehrig2024low}.
Interestingly, a smaller model does not yield lower latency in our setting (\eg on Gen1, \(3.43\,\mu s\) for SSLA-S and  \(2.44\,\mu s\) for SSLA-B), because the actual runtime also depends on hardware factors such as vectorization efficiency and memory access patterns.
The SSLA module has a constant per-event inference FLOPS of \(\mathcal{O}(P^2D^2_{out})\), making the latency independent of resolution.
While CPU does not fully translate our FLOPS efficiency into latency gains \cite{gehrig2024low}, further runtime latency reduction could be achieved on specific hardware, such as FPGAs  \cite{jeziorek2026hardware} or neuromorphic accelerators \cite{zhang2026compute}.  

% table 4

\begin{table}[htbp]
\centering
\caption{\textbf{Inference latency.} Latency refers to the time used for the recurrent model to process a new event.}
\label{tab:timing-inference}
\begin{tabular}{lccccc}
\toprule
\multirow{2}{*}{Dataset}& \multirow{2}{*}{Resolution} & \multicolumn{4}{c}{Latency (\(\mu s\))} \\
\cmidrule(lr){3-6}
& & SSLA-S & SSLA-B & SSLA-M & SSLA-L \\
\midrule
Gen1 & 304 \(\times\) 240 & 3.43 & 2.44 & 6.02 & 7.20  \\
N-Caltech101 & 240 \(\times \) 180 & 3.60 & 2.61 & 6.50 & 8.01 \\
\bottomrule
\end{tabular}
\end{table}

\subsection{Ablation Study}

\subsubsection{Efficiency Attribution.} 
To isolate the sources of efficiency in SSLA-Det, we compare SSLA-S with three variants: (i) removing temporal dropout (TD), (ii) further replacing the SSLA module with a dense-activation counterpart that retains the MOS decomposition and PAP but activates all patches per event, and (iii) removing sparse pooling (SP). 
As shown in \cref{tab:efficiency-attribution}, the SSLA module contributes the dominant computational cost reduction (380 \(\times\)) at no accuracy cost, while TD provides an additional 10 \(\times\) reduction as an accuracy-efficiency trade-off. 
SP does not change the per-event FLOPS since it only downscales the coordinates of events without reducing the event count, but is essential for detection.

\begin{table}[htbp]
\centering
\caption{\textbf{Efficiency Attribution.} We isolate the contribution of SSLA-Det components on the validation set of Gen1. TD refers to temporal dropout, SP refers to sparse pooling, and we remove SSLA by changing it to its dense counterpart and keeping the MOS and PAP designs.}
\label{tab:efficiency-attribution}
\begin{tabular}{lcccc}
\toprule
Configuration & SSLA-S & w/o TD & w/o SSLA \& TD & w/o SP \\
\midrule
MFLOPS/ev $(\downarrow)$ & 0.102 & 1.02 & 388 & 0.102 \\
mAP $(\uparrow)$ & 0.335 & 0.370 & 0.370 & 0.014 \\
\bottomrule
\end{tabular}
\end{table}

\subsubsection{Effect of Spatial Sparsity.}

We replace the SSLA module with a standard linear attention (LRU).
We use \(D_{out}\) of the first stage 12 and 36, to keep same \(D_{out}\) and similar FLOPS as SSLA-S.
\cref{tab:ablation-ssla} shows that using a standard linear attention fails in both cases, which is considered mainly due to the lack of fine-grained state.
In particular, SSLA-S maintains a state 380 \(\times\) larger than LA (\(D_{out}=36\)) with similar FLOPS (0.102 M/ev vs. 0.093 M/ev).
This demonstrates the importance of state-level sparsity.

% Table 5
\begin{table}[htbp]
\centering
\caption{\textbf{Effect of Spatial Sparsity.} We use linear attention (LA) with same embedding dimension (first stage \(D_{out}=12\)) and similar FLOPS (first stage \(D_{out}=36\)) compared to SSLA-S. We report accuracy on the validation set of Gen1. State refers to the size of hidden state in the last layer, which reflects the capability to model fine-grained spatial representations.}
\label{tab:ablation-ssla}
\begin{tabular}{lccccccc}
\toprule
Model & mAP\((\uparrow)\) & AP\(_{50}\)\((\uparrow)\) & AP\(_{75}\)\((\uparrow)\) & MFLOPS/ev\((\downarrow)\) & Params (M) & State (K) \\
\midrule
SSLA-S              & \textbf{0.335} & \textbf{0.610} & \textbf{0.322} & 0.102 & 0.508 & 106.9\\
LA (\(D_{out}=12\)) & 0.001 & 0.004 & 0.000 & \textbf{0.011} & 0.130 & 0.094 \\
LA (\(D_{out}=36\)) & 0.001 & 0.003 & 0.000 & 0.093 & 1.20  & 0.281 \\
\bottomrule
\end{tabular}
\end{table}

\subsubsection{Effect of Position-Aware Projection.}
To ablate PAP, we replace it with a position-irrelevant learnable linear projection.
The results are summarized in \cref{tab:ablation-pos}.
Removing either input or output PAP causes a significant accuracy drop, and removing both results in catastrophic failure, with the mAP collapsing to only 0.014, which highlights its importance for encoding spatial priors in the SSLA module.

% Table 6

\begin{table}[htbp]
\centering
\caption{\textbf{Effect of Position-Aware Projection.} Input and Output refer to the PAP with \(\mathbf{W}^{\text{in}}\) and \(\mathbf{W}^{\text{out}}\), respectively. We report accuracy on the validation set of Gen1.}
\label{tab:ablation-pos}
\begin{tabular}{cccccc}
\toprule
\multicolumn{2}{c}{Position-Aware Projection} & \multirow{2}{*}{mAP\((\uparrow)\)} & \multirow{2}{*}{AP\(_{50}\)\((\uparrow)\)} & \multirow{2}{*}{AP\(_{75}\)\((\uparrow)\)} & \multirow{2}{*}{MFLOPS/ev\((\downarrow)\)} \\
\cmidrule(lr){1-2}
\multicolumn{1}{c}{Input} & \multicolumn{1}{c}{Output} & & & & \\
\midrule
\checkmark & \checkmark & \textbf{0.335} & \textbf{0.610} & \textbf{0.322} & 0.102 \\
\checkmark &            & 0.306          & 0.582 & 0.280 & 0.085 \\
           & \checkmark & 0.224          & 0.473 & 0.184 & 0.089 \\
           &            & 0.014          & 0.048 & 0.006 & \textbf{0.072} \\
\bottomrule
\end{tabular}
\end{table}

\subsubsection{Effect of Patch Size.}
\(P\) controls the receptive field of event interaction.
As shown in \cref{tab:ablation-patch}, a smaller patch size (\(P=2\)) reduces FLOPS (0.047 M/ev) but leads to a mAP drop (0.200).
Conversely, \(P=4\) boosts the mAP to 0.371 but increases the FLOPS to 0.179 M/ev, showing an accuracy-efficiency trade-off.
Besides, for training, increasing \(P\) results in higher GPU memory consumption and longer training time.
Therefore, we select \(P=3\) as our default configuration as it yields a reasonable trade-off between efficiency and accuracy.

\begin{table}[htbp]
\centering
\caption{\textbf{Effect of Patch Size.} We report accuracy on the validation set of Gen1.}
\label{tab:ablation-patch}
\begin{tabular}{ccccc}
\toprule
Patch Size & mAP & AP\(_{50}\) & AP\(_{75}\) & MFLOPS/ev \\
\midrule
\(P=2\) & 0.200          & 0.440          & 0.150 & \textbf{0.047} \\
\(P=3\) & 0.335          & 0.610          & 0.322          & 0.102 \\
\(P=4\) & \textbf{0.371} & \textbf{0.656} & \textbf{0.364} & 0.179 \\
\bottomrule
\end{tabular}
\end{table}

\section{Limitation and Discussion}

This work focuses on event-based low-latency object detection.
While hybrid event-image models have become a recent research trend \cite{gehrig2024low, li2025asynchronous}, they also introduce additional challenges, including event-image alignment, extra sensor requirements, and high system complexity.
Besides, in principle, our method is also compatible with the hybrid framework, since image features from dense models can be injected into the intermediate layers of our model.
Exploring this event-image fusion in SSLA is an interesting direction for our future work.

Although SSLA-Det achieves SOTA performance among asynchronous event-based object detection methods, a gap remains compared with synchronous methods. For example, as shown in \cref{tab:gen1-results} on Gen1, SSLA-L has an mAP of 0.375, whereas synchronous SOTA methods exceed 0.5 mAP.
This gap is expected, as asynchronous and synchronous methods target fundamentally different objectives along the accuracy-efficiency trade-off and are not directly comparable. 
Synchronous methods accumulate events into image-like representations and perform dense image-level inference, which allows for more information aggregation, the use of image-based neural network architectures and pretrained weights \cite{zubic2023chaos, hao2026maximizing}, and models with larger parameter count \cite{zubic2023chaos, fan2025eventpillars}, but at the cost of larger computational cost (\cref{tab:gen1-results}) and millisecond-level latency. 
Asynchronous models, in contrast, aim to realize the low-latency advantage of event cameras at the neural network level, giving predictions event-by-event at minimal latency. 
This structurally constrains parameter count, information aggregation, and architectural choices, naturally limiting accuracy. 
Therefore, the remaining accuracy gap should be understood as part of the accuracy-latency trade-off in low-latency event-based perception, rather than a methodological shortcoming.
Further improving this trade-off while preserving \(\mu\)s-level per-event latency remains an important direction for future work.

\section{Conclusion}

In this paper, we propose SSLA, a novel linear attention module with spatial sparsity and efficient parallel training capability for event sequence modeling.
We develop SSLA-Det, the first end-to-end asynchronous linear attention-based model for event-based object detection.
Experimental results on Gen1 and N-Caltech101 show that SSLA-Det achieves SOTA asynchronous accuracy with significantly lower FLOPS than previous asynchronous baselines. 
We believe that SSLA provides a promising direction for low-latency, high-performance event-based perception.

\section*{Acknowledgements}
This work was supported by the State Key Laboratory of Precision Measurement Technology and Instruments (2025PMTI03), and STI 2030-Major Projects (2021ZD0200300).

% ---- Bibliography ----
%
% BibTeX users should specify bibliography style 'splncs04'.
% References will then be sorted and formatted in the correct style.
%

\bibliographystyle{splncs04}
\bibliography{main}
\end{document}